\documentclass[10pt,twocolumn,letterpaper]{article}

\usepackage{cvpr}              
\usepackage[accsupp]{axessibility}  

\usepackage{graphicx}
\usepackage{svg}
\usepackage{capt-of} 

\usepackage{algorithm}
\usepackage{algpseudocode}
\usepackage{amsmath}

\usepackage{listings,newtxtt}
\definecolor{codegreen}{rgb}{0,0.6,0}
\definecolor{codegray}{rgb}{0.5,0.5,0.5}
\definecolor{codepurple}{rgb}{0.58,0,0.82}
\definecolor{backcolour}{rgb}{0.95,0.95,0.92}

\newcommand{\sysname}{\textsc{\uppercase{GRS}}}

\definecolor{cvprblue}{rgb}{0.21,0.49,0.74}
\usepackage[pagebackref,breaklinks,colorlinks,allcolors=cvprblue]{hyperref}

\def\paperID{0005} 
\def\confName{CVPR}
\def\confYear{2025}

\title{\LARGE \bf
GRS: \textit{G}enerating \textit{R}obotic \textit{S}imulation Tasks from Real-World Images}

\author{Alex Zook$^{1}$, Fan-Yun Sun$^{2}$, Josef Spjut$^{1}$, Valts Blukis$^{1}$, Stan Birchfield$^{1}$, Jonathan Tremblay$^{1}$ \\
$^{1}$NVIDIA $^{2}$Stanford University \\
{\tt\small \{azook, jspjut, vblukis, sbirchfield, jtremblay\}@nvidia.com} \\
{\tt\small sunfanyun@cs.stanford.edu}
}

\begin{document}

\maketitle

\begin{abstract}
We introduce GRS (Generating Robotic Simulation tasks), a system addressing real-to-sim for robotic simulations.
GRS creates digital twin simulations from single RGB-D observations with solvable tasks for virtual agent training.
Using vision-language models (VLMs), our pipeline operates in three stages: 1)~scene comprehension with SAM2 for segmentation and object description, 2)~matching objects with simulation-ready assets, and 3)~generating appropriate tasks.
We ensure simulation-task alignment through generated test suites and introduce a router that iteratively refines both simulation and test code.
Experiments demonstrate our system's effectiveness in object correspondence and task environment generation through our novel router mechanism.
\end{abstract}

\section{INTRODUCTION}

Digital twin simulations are valuable in game generation, AR/VR, robotics, and human training simulations. 
The real-to-sim problem of creating digital twins from real-world observations involves three steps: 1)~understanding the scene, 2)~finding/creating assets to populate the scene, and 3)~generating tasks for virtual agents.
Beyond robotics applications, digital twins are relevant to educational games and training simulations that mirror real environments while maintaining interactive, task-oriented gameplay.

Creating simulations from real-world observations requires decomposing scenes into their components, spatial relationships, and visual properties.
Existing methods include 3D reconstruction~\cite{Jiang_2022_CVPR}, manipulating latent spaces~\cite{paxton2019visual,black2023zero}, pose estimation~\cite{deitke2023phone2proc}, and inverse rendering~\cite{kulits2024re,guo2024physically}.
Recent research has leveraged LLMs to generate tasks~\cite{wang2023gensim,zeng2024learning,chen2024roboscript} or scenes~\cite{wang2023robogen,goddard2018molecular,yang2024holodeck}.

\begin{figure}
    \centering
    \includegraphics[width=0.99\linewidth,trim={0 0 235px 0}, clip]{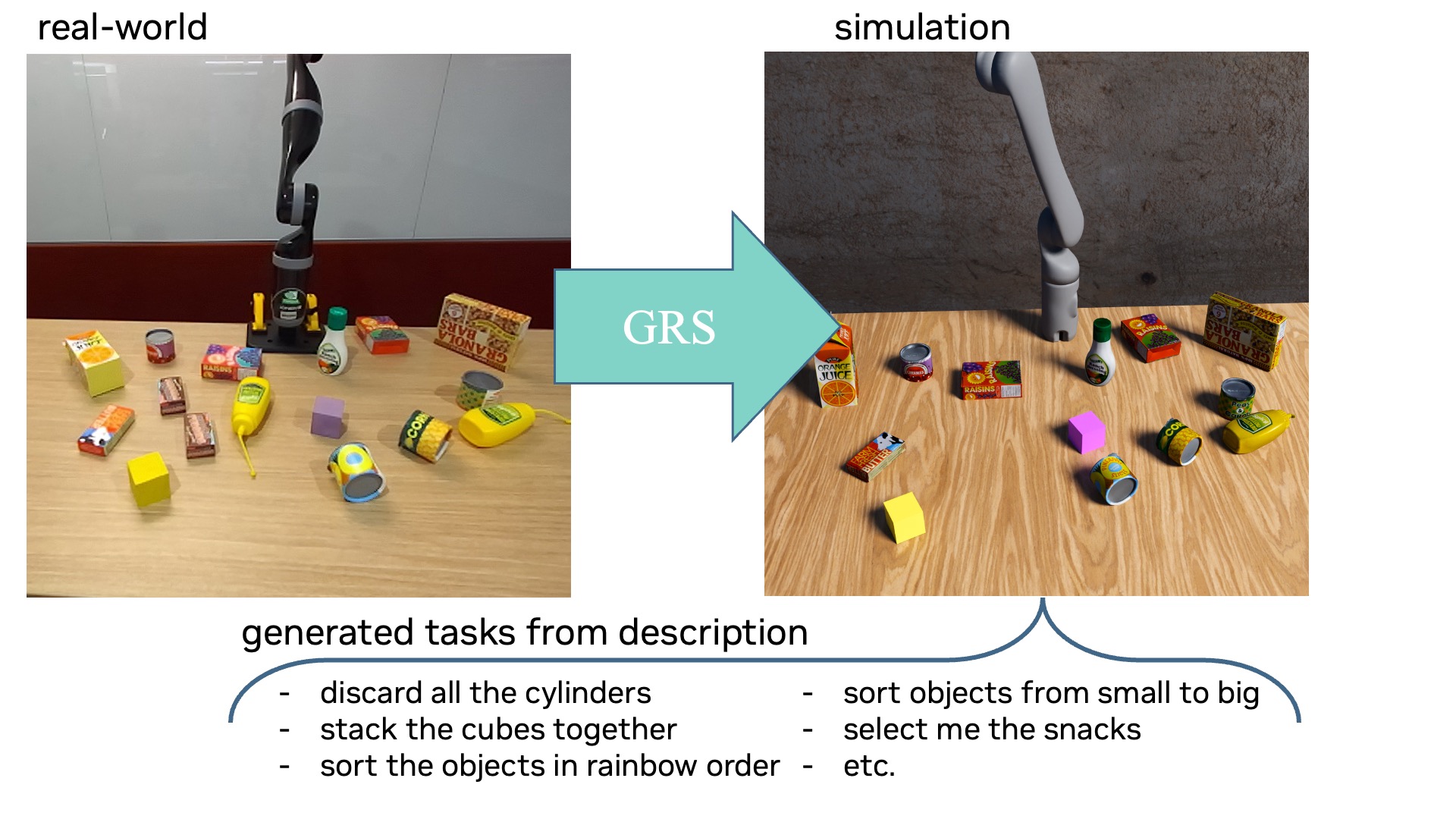}
    \caption{
        GRS solves the problem of generating robotics simulations with solvable tasks from real-world images. 
        During task generation, GRS can use a subset of objects and change
        the object orientation and positioning to provide interesting variations on the initial scene.
    }
    \label{fig:problem}
\end{figure}

We present \sysname, a system that extracts scene descriptions from single real-world observations and generates diverse, solvable tasks (Figure~\ref{fig:problem}). 
Our system segments objects using SAM2~\cite{ravi2024sam}, processes each part with a VLM to generate text descriptions, matches these against simulation-ready assets, and creates contextually appropriate tasks for the digital twin world.

\sysname{} introduces a novel approach to maximize alignment between task specifications and simulation creation.
The core of our innovation lies in two key components: 
First, we leverage a dual-generation process where each simulation is accompanied by a tailored set of tests.
These tests are specifically generated to evaluate the simulation's fidelity to the original task description, ensuring a comprehensive validation process. 
Second, we introduce an LLM-based \textit{router} system that analyzes simulation performance, including runtime errors and test outcomes. 
This router makes informed decisions on whether to refine the simulation or adjust the test battery, optimizing the alignment process. 
Through iterative refinement driven by an LLM, \sysname{} generates highly robust simulations that accomplish intended tasks with accuracy and reliability. 

In our experiments, we demonstrate that \sysname{} can succeed at the real-to-sim task with a single RGB-D observation.
First, we show that our scene object identification process is highly accurate and provide thorough ablations of alternative methods for visual object correspondence.
Second, we demonstrate the value of VLMs for improving the generation of tasks that more closely match their input environments.
Third, we show that our router improves the rate of generating simulations effective for robot policies compared to baseline methods that only do code repair.
We make the following contributions:
1) a novel system that enables real-to-sim for simulation generation, and
2) demonstrating the efficiency of the method on a battery of real and simulated tests.

\section{RELATED WORK}

\noindent\textbf{Scene Understanding} is a crucial component in generating robotic simulation tasks from images. 
Various computer vision systems have been developed to tackle this challenge.
The Segment Anything Model (SAM) has emerged as a powerful tool for object segmentation in images. 
This model can identify and segment objects in a scene with high accuracy, providing a foundation for further scene analysis~\cite{ravi2024sam}.
Additionally, open-world localization models like OWL-ViT have shown promise in adapting pretrained open-world image models to video tasks~\cite{minderer2022simple}.
Compared to our approach, special text has to be included to ground the objects which could lead to missing objects.
By contrast, our method leverages the large knowledge of VLMs to describe objects from which we can deduce if a robot can manipulate them or not. 
Phone2Proc~\cite{deitke2023phone2proc} uses an API to generate a well defined 3d interior, and then procedurally place assets. 
In addition, recent advances in 3d scene retrieval have focused on using LLMs for visual program synthesis~\cite{chen2023visual,ellis2018learning,tam2024scenemotifcoder} or 3d scene generation~\cite{aguina2024open,feng2024layoutgpt,huang2024blenderalchemy}.

\vspace{1mm}
\noindent\textbf{Simulation Creation} with LLMs has been explored to automate and enhance the creation of simulations for various applications~\cite{zeng2023agenttuning,hu2024agentgen,sun2024factorsim}. 
GenSim~\cite{wang2023gensim} uses LLMs to generate robotic simulation tasks, demonstrating the potential of LLMs to create diverse and complex simulation scenarios. 
Zeng \textit{et al.}~\cite{zeng2024learning} present a system to generate reward functions based on a task definition. 
RoboGen~\cite{wang2023robogen} presents a generative robotic agent that automatically learns diverse robotic skills at scale via generative simulation.
This system leverages foundation and generative models to create diverse tasks, scenes, and training supervision.
Similar to our method, the system generates tasks using the scene information and also proposes a training approach based on the type of task being generated. 
Holodeck~\cite{yang2024holodeck} offers language-guided generation of 3D embodied AI environments. 
This approach shows the potential of integrating natural language processing with 3D environment creation for robotic simulations. 
Compared to our method, these methods do not ensure the tasks are representative of the task definition, or scenes they generate are solvable or functional.
FactorSim~\cite{sun2024factorsim} uses ideas from factorized Markov Decision Processes to generate games and trains agents using reinforcement learning in those games.
Our router extends the ideas in this work to improve automated testing and iteration on game code.

World models have emerged as a way to create interactive simulations.
Diffusion models can be trained on data from existing games to provide interactive game engines with new controls through text or image prompting.
Examples include Genie~\cite{bruce2024genie,google2024genie2}, WHAM~\cite{kanervisto2025wham}, DIAMOND~\cite{alonso2024diamond}, GameNGen~\cite{valevski2024gamengen}, and Oasis~\cite{oasis2025oasis}.
Another approach is to tune video-based generative models to condition on control input for actions or camera manipulation.
Examples include Promptable Game Models~\cite{menapace2024promptable}, The Matrix~\cite{feng2024matrix}, GameGen-X~\cite{che2024gamegenx}, and GameFactory~\cite{yu2025gamefactory}.
Unlike these methods we produce code readily uses existing game engines as simulators, complementing existing workflows and ensuring hard physical constraints are enforced by the simulator.
These are crucial features for domains where real-world physics and behavior matter, as when games are used to train humans for real-world tasks like surgery or piloting.

\section{METHOD}

Our approach to real-to-sim task generation has two phases: 1)~scene comprehension, and 2)~simulation generation and evaluation.
Initially, we process an input RGB-D image to extract scene information, including bounding boxes and segmentation masks.
Subsequently, we establish correspondences between these extracted elements and simulation-ready assets.

Using this scene data, we formulate a task for a robotic system to execute. 
The extracted 3D assets and scene information are the key inputs for generating a simulation program and associated test cases.
We introduce a novel iterative refinement process, termed \textit{router}, which iteratively enhances both the simulation program and test cases until a policy successfully completes the prescribed task.
Following GenSim~\cite{wang2023gensim}, \textit{tasks} are the text description of the goals and/or actions to be executed by a robotic system, and \textit{simulations} are the code that implements the task.
This distinction separates conceptual instructions (task) and their concrete implementation (simulation) in our framework.

While our primary focus is on robotic applications, our methodology is directly applicable to video game development.
In game development, the same challenges arise: converting real environments to interactive virtual spaces with meaningful objectives.
Our approach to generating solvable tasks can be used for creating game levels with appropriate difficulty, generating game mechanics tied to physical objects, and ensuring player objectives are achievable---critical aspects of game design.

\subsection{Scene Comprehension}
\label{sec:scene-comprehension}

\begin{figure}
    \centering
    \includegraphics[width=0.99\linewidth,trim={0 0 435px 0}, clip]{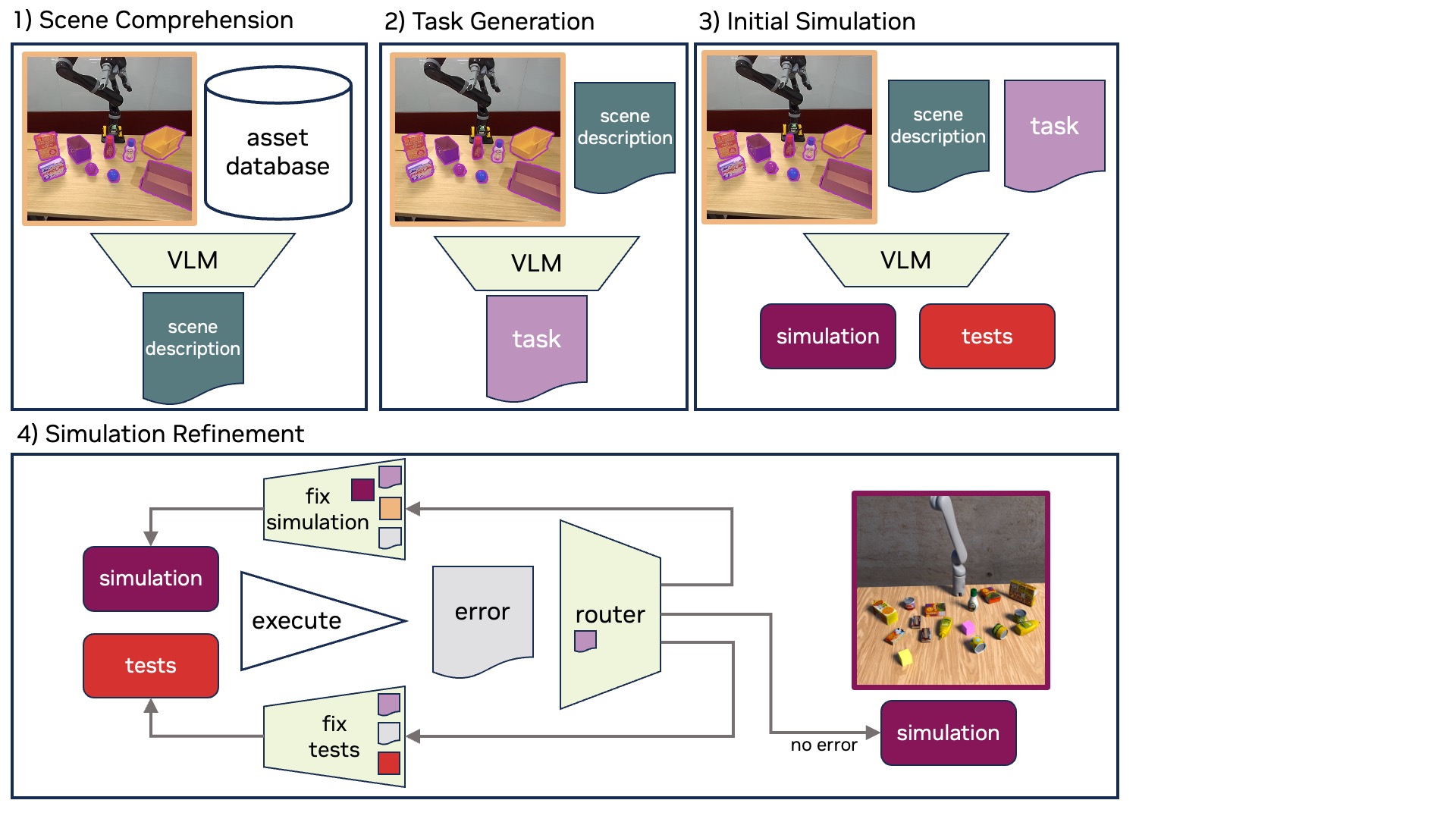}
    \caption{
\sysname{} workflow for real-to-sim conversion. The process has four stages: 1)~scene description generation using segmented images and simulation-ready assets; 2)~task creation based on the scene; 3)~initial simulation and test code generation; 4)~iterative refinement until error-free simulation is achieved. Color and shape coding is used for certain inputs to enhance visual clarity.
    }
    \label{fig:vision}
\end{figure}

Scene description follows a two-stage approach of image segmentation followed by image description, as illustrated in the first entry in Figure~\ref{fig:vision}.
This divide-and-conquer approach ensures a detailed understanding of the scene, facilitating accurate simulation and task generation.

\noindent\textbf{Image Segmentation.} We use SAM2~\cite{ravi2024sam} to segment the input image into crops, which often results in over-segmentation of object parts and background elements.
This granular detail provides a foundation for nuanced scene understanding.
For simulation use, we map each crop to a 3D bounding box by using depth data to transform segmented pixels to 3D positions in the robot's coordinate frame, then fitting them within axis-aligned bounding boxes.

For later use in scene generation, we map each image crop to a 3D bounding box of the candidate object.
For each crop the segmented pixels are mapped to their corresponding 3D positions using depth data.
The 3D coordinates are then transformed into the robot's coordinate frame through a calibrated transformation matrix to spatially align with the robot's environment. 
Once in the robot frame, the candidate object's position and extent are fitted within an axis-aligned 3D bounding box, enabling reliable geometric matching.

\noindent\textbf{Object Correspondence.} The object correspondence process aims to link candidate objects with appropriate 3D assets for simulation.
Our approach involves three steps:
1)~\textit{Asset Database Creation:} We create a database of 3D asset descriptions by prompting a VLM to analyze multiple renders of each asset.
This process generates rich, multi-perspective descriptions of each 3D object in our asset library.
We perform this pre-processing step once and retain the text description database for reuse when evaluating different target scenes.
2)~\textit{Candidate Object Description:} We use the same VLM to describe the candidate object crops derived from our segmentation process.
This description is based solely on their visual information, ensuring a consistent basis of comparison with the asset database.
This step occurs once per real-to-sim target scene, as each scene has a different set of cropped images as output from the image segmentation.
3)~\textit{Description Comparison:} We use the VLM to compare the candidate object text description and the cropped real image to the descriptions in the asset database.
This matches each candidate object to a 3D asset in the asset database or identifies that there is no object in the cropped image (to address over-segmentation).
This step is also performed once per target scene.

The result of this process is a comprehensive list of scene assets, each associated with its specific bounding box information obtained during the initial image analysis.
This detailed mapping forms the foundation for accurate scene reconstruction in the simulation environment.

\subsection{Simulation Generation and Evaluation} 

The challenge of simulation generation lies in translating real-world objectives into a simulator-compatible program for robot execution.
This code must precisely define the simulator's starting configuration and desired end state.
Crucially, the generated simulation should run without errors and be optimized for feasibility, allowing a robot policy to complete it successfully within an acceptable time frame.

\sysname{}'s simulation generation process takes as input a scene image and a scene description, summarized in Algorithm~\ref{alg:task-generation}. 
Inspired by GenSim~\cite{wang2023gensim}, we divide the simulation generation into two phases: 1) genearing an abstract task definition and selecting appropriate scene assets, and 2) writing the simulation program for the task.
Our approach enhances both steps by incorporating scene images and using a VLM for input processing, departing from GenSim's LLM-based method.
Unlike GenSim, we do not use predefined assets; instead, we leverage candidate assets and their placements identified during object correspondence.
This allows our task generation to benefit from both the visual context of the scene and the textual descriptions of available assets.

\noindent\textbf{Task Definition Generation.} Given the scene descriptions and the selected visual assets our system first generates a task definition, depicted in the second entry of Figure~\ref{fig:vision}.
As input we provide the scene information, image, and asset descriptions, prompting for creation of a contextually relevant robotics task.
To accommodate a variety of potential tasks we allow for the task to use a subset of the observed assets.
Our focus is on generating tasks that are both practical and challenging for robotic manipulation systems.
These tasks typically involve manipulating objects within the scene in specific ways, such as stacking particular items or grouping objects by category.
Listing~\ref{lst:task-def} shows an example of a task definition generated by our system, where the LLM chose to make the task packing food items in a box.

This approach allows for the creation of a wide range of tasks, from simple object manipulation to more complex spatial reasoning and organizational challenges, all tailored to the specific objects and layout present in the given scene.
By leveraging the detailed scene understanding achieved through our segmentation and object correspondence processes, we ensure that the generated tasks are not only diverse but contextually appropriate to the real scene and feasible within the simulated environment.

\begin{lstlisting}[label=lst:task-def, caption=Example task definition., float=b]
"task-name": "pack-food-items-in-box",
"task-description": "Pick up all the food items and place them inside the open box.",
"assets-used": [
    "HOPE/GranolaBars.urdf",
    "HOPE/MacaroniAndCheese.urdf",
    "HOPE/OrangeJuice.urdf",
    "HOPE/Raisins.urdf",
    "HOPE/Mayo.urdf",
    "HOPE/Mustard.urdf",
    "HOPE/Cookies.urdf",
    "container/container-template.urdf"
]
\end{lstlisting}

\begin{algorithm}[tb]
\caption{Simulation Generation Algorithm}
\label{alg:task-generation}

\begin{algorithmic}[1]
\Procedure{TaskGeneration}{}
  \State \textbf{Inputs:} image, scene description
  \State \textbf{Outputs:} simulation, tests
  \State simulation, tests $\gets$ VLM$($image, scene description$)$

  \Repeat
    \State error $\gets$ Evaluate$($simulation, tests$)$
    \If{error $\neq \emptyset$ }
    \State route based on error:
      \State a) fix simulation, or
      \State b) fix tests
    \EndIf
  \Until{error $= \emptyset$}
  \State \textbf{Return:} simulation, tests
\EndProcedure
\end{algorithmic}
\end{algorithm}

\noindent\textbf{Simulation Program Generation.} We next generate the simulation code using a VLM that is provided the scene image, task definition, and asset descriptions, shown in the third entry in Figure~\ref{fig:vision}.
Following GenSim, the simulation code is python code that subclasses a generic Task class.
The LLM prompt includes the object bounding box positions as floats and strings referencing the URDF assets to load in the simulator.
The LLM is permitted to modify the object list and positions during iteration on the task.

To ensure the generated simulation is valid for the robot task, we also generate a battery of tests intended to ensure the task can be completed by a robot policy.
The test program is generated by providing the simulation program and the task description as input to an LLM.
Tests are implemented as python unit tests that use the `unittest` library.
Listing~\ref{lst:task-code} in the appendix shows the final simulation output by our system for the task definition in Listing~\ref{lst:task-def}.

To align the task description and the generated simulation, we introduce a novel LLM \textit{router} system that dynamically iterates over the simulation program and tests. 
The algorithm follows a straightforward yet powerful approach:
1)~\textit{Run Tests}: Execute tests on the simulation program and collect any errors.
2)~\textit{Router}: Use the task description and error information to determine whether to update the generated test program or the simulation program.
3)~\textit{Fix}: Revise the appropriate components using either a VLM for simulation code or an LLM for test code, considering inputs such as scene image, errors, and task definition.
4)~Repeat this cycle until execution occurs without errors.
This algorithm is visually illustrated as the last entry in Figure~\ref{fig:vision}. 
This process is simple, yet effective, enabling the system to reason about the components of simulation generation and their relationships.
By refining both the simulation and its associated tests using the task definition as guidance, our \textit{router} ensures alignment between the conceptual task description and its practical implementation in the simulated environment.
Listing~\ref{lst:test-code} shows one of the tests generated by our system and Listing~\ref{lst:router} shows the prompt for the router.

\begin{lstlisting}[language=Python, label=lst:test-code, caption=Example test code., float=tb]
def test_oracle_picks_and_places_objects(self):
    oracle_agent = self.task.oracle(self.env)
    obs = self.env.reset()
    info = self.env.info

    pick_count = 0
    place_count = 0

    for _ in range(self.task.max_steps):
        act = oracle_agent.act(obs, info)
        if act is None:
            break  # Ensure we handle cases where the oracle cannot generate an action
        obs, reward, done, info = self.env.step(act)
        
        if act and 'pose0' in act and 'pose1' in act:
            pick_count += 1
            place_count += 1
        
        if done:
            break
    
    # Ensure that the oracle agent attempted to pick and place objects.
    self.assertGreater(pick_count, 0, "Oracle agent did not attempt to pick any objects.")
    self.assertGreater(place_count, 0, "Oracle agent did not attempt to place any objects.")
\end{lstlisting}

\begin{lstlisting}[language=Python, label=lst:router, caption=Prompt for router., float=b]
You are an AI in robot simulation code and task design.
My goal is to design diverse and feasible tasks for tabletop manipulation.

Below are the results of running the current unit tests on the simulation code for a task.

Should I next attempt to fix the task code or fix the unit test code?

Provide a detailed explanation of what is going wrong according to the test results.
Then describe what would be the easiest fix to make to reduce the number of failing tests.
Consider whether it would be better to modify the tests to get more detailed errors or to modify the code to address the current errors.

End your reply with directions on whether to change the task code or tests.
Only provide one of the following values as a valid output, enclosed in a code block (```):
- "fix_code"  # fix the task code
- "fix_test"  # fix the unit tests


Here is the task definition:
{task}

Here are the unit test results:
{test_results}
\end{lstlisting}

We focus on robotic simulations suitable for policy execution or training.
The LLM is prompted to write tests to ensure an oracle robot policy can succeed at the task.
The includes API information for initializing a generic task in the simulator and calling an oracle agent in the simulator, along with a simplified execution loop for environment observation and action.
Successful execution by an oracle agent is a stringent but valuable criterion, requiring error-free code that specifies achievable objectives within the simulator's physical constraints. 
While an alternative approach could involve unit tests that only check scene definition validity, we opt for testing with the oracle robot policy. 
Testing with an oracle incurs greater simulation generation and validation costs, but increases the likelihood of successful downstream task generation.
By using an LLM to author the tests using an oracle we ensure the task details represent a feasible task for downstream applications training agents in the simulator.

\section{EXPERIMENTS}

Our method aims to improve asset retrieval accuracy and improves task-simulation alignment. 
Given the scarcity of benchmarks for real-to-sim translation, we introduce a table-top robotics-inspired task of 10 distinct scenes with an average of 15 objects each.
Objects are from the HOPE dataset~\cite{tyree20226} of common grocery items with available 3D models, supplemented with colored cubes and containers.

For each scene, we recorded a 1080p RGB image and point cloud using a ZED 2 camera.
Object correspondence generated a scene description defining objects with their crops, bounding boxes, and text descriptions.
In experiments the a given robot is placed at a default center position in the scene, mirroring our scenario of converting real-world robot setup images to simulations (Figure~\ref{fig:task_qual}).

We performed three experiments: (1)~evaluating our object correspondence methodology, (2)~evaluating of our task generation pipeline, and (3)~demonstrating scalability to more complex scenes with larger asset databases.
Results show that VLMs using images and text descriptions yield the highest accuracy for object correspondence, our simulation generation has greater efficiency and performance compared to previous methods, and our approach has potential to scale to more complex scenes with larger asset databases.

\subsection{Object Correspondence}

Our object correspondence experiment assessed a model's ability to retrieve the correct assets for a captured scene.
For each 3D model in our dataset, we generated three rendered views of the object by randomly positioning the camera around the object in an empty space while maintaining focus on it.
We then had a VLM produce a detailed object description based on these renders, encompassing features like shape, colors, branding, or patterns.
A VLM also generated text descriptions for each segmented part of the real-world scene image, see Section~\ref{sec:scene-comprehension} for further details on object description generation and object correspondence.

This setup allowed us to evaluate three object correspondence methods: 
1)~matching image text descriptions to asset descriptions ({\tt text}), 
2)~matching images to asset descriptions ({\tt image}), and 
3)~matching both images and text descriptions with asset descriptions ({\tt ours}). 
For each method, we tested both GPT4o~\cite{achiam2023gpt4} and Claude-3.5-Sonnet~\cite{anthropic2024claude}.
For object correspondence the VLM was prompted to pick the best object description from the set of generated descriptions, or indicate that no object was present.

Table~\ref{tab:vlm_label_summary_10scenes} presents the object retrieval results.
We used F1 scores as a balanced measure of precision and recall (higher scores indicate better performance), aggregating over 10 runs of each task, where each run produced asset descriptions and text descriptions (where appropriate) before performing object correspondence.
GPT4o demonstrated superior performance in all tasks.

Additionally, we included a baseline using CLIP embedding distance~\cite{radford2021learning}.
We embedded the images of the assets in the database and compared the average of the embedding of the three images of an asset to an embedding of the cropped image to find the matching asset with the smallest CLIP embedding distance.
If a real-world image crop had a CLIP similarity score below 0.5 with all assets in the scene, it was considered ``not an object.''

Our approach outperformed both baselines as simply matching assets based on CLIP embedding distance is not robust to occlusions and variations in object poses and lighting conditions.
We performed Kruskal-Wallis testing for significance since the underlying data violated the normality assumption that would be required for ANOVA testing.
The Kruskal-Wallis test found statistically significant ($p<0.05$) differences for task type, model, and their interaction, indicating our results are statistically significant differences.

\begin{table}[tb]
    \caption{Performance of different models on each task as F1 mean with 95\% confidence interval (in parentheses). Highest mean in bold. }
    \begin{center}

\begin{tabular}{llll}
\hline
task type & model & F1 (95\% CI) \\
\hline
CLIP & -- & 0.76 (0.63,0.88) \\
{\tt text} & Claude-3.5-Sonnet & 0.67 (0.65,0.69) \\
{\tt text} &  GPT4o & 0.83 (0.81,0.85) \\
{\tt image} & Claude-3.5-Sonnet & 0.55 (0.53,0.56) \\
{\tt image} & GPT4o  & 0.88 (0.87,0.89) \\
{\tt ours} &  Claude-3.5-Sonnet & 0.62 (0.59,0.64) \\
{\tt ours} &  GPT4o & \textbf{0.89} (0.87,0.90) \\
\hline
\end{tabular}

    \end{center}

    \label{tab:vlm_label_summary_10scenes}
\end{table}

\subsection{Robot Task Generation}

\begin{figure}
    \centering
    \includegraphics[width=1.2\linewidth, clip]{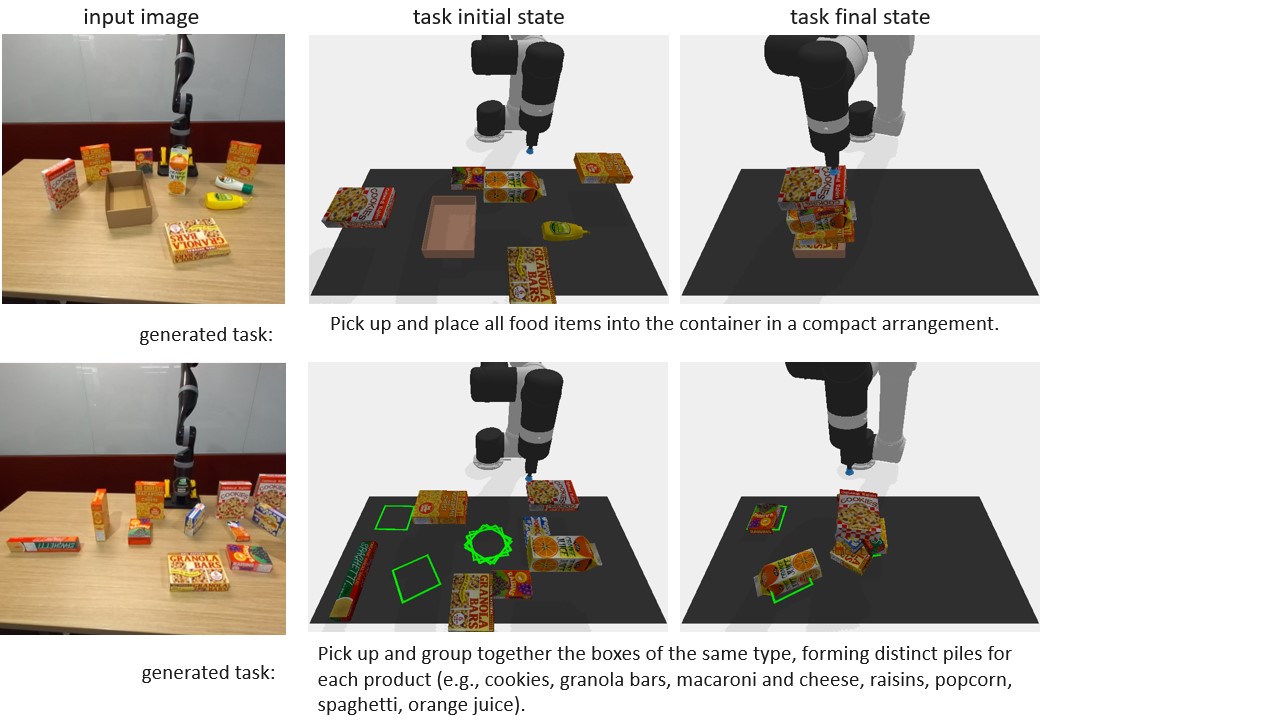}
    \includegraphics[width=1.2\linewidth, clip]{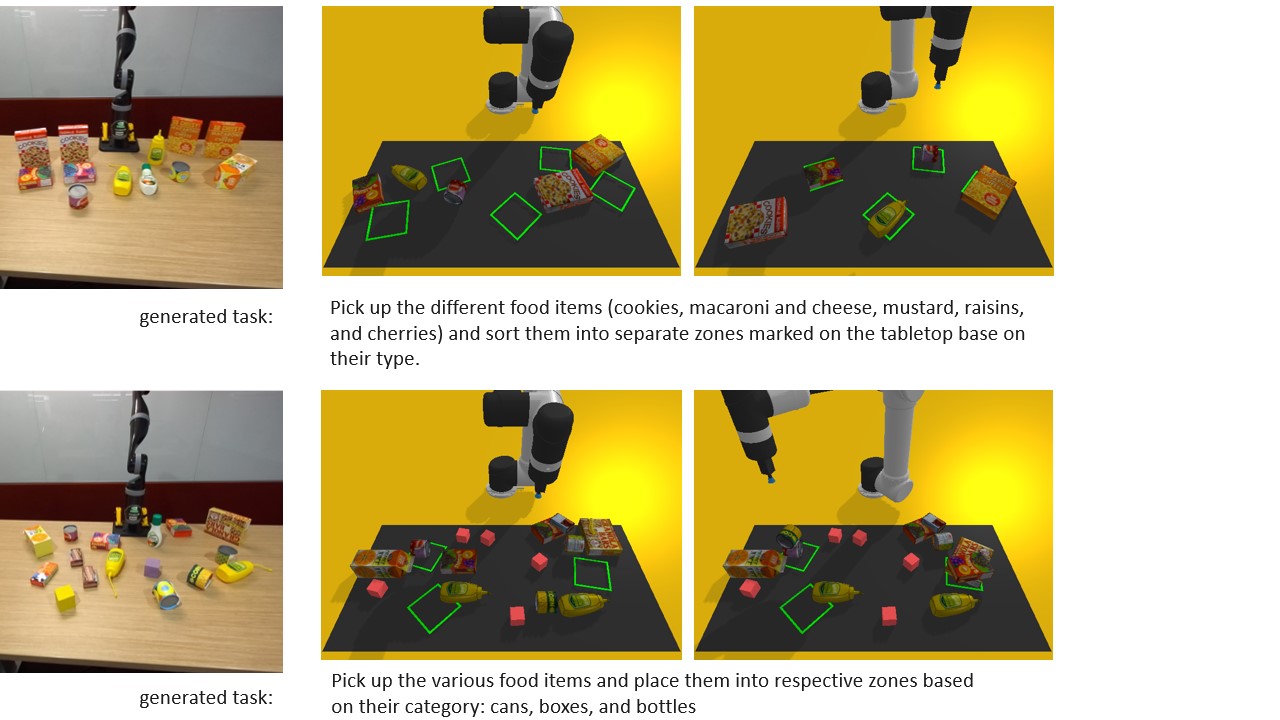}
    \includegraphics[width=1.2\linewidth, trim={0 280px 0 0}, clip]{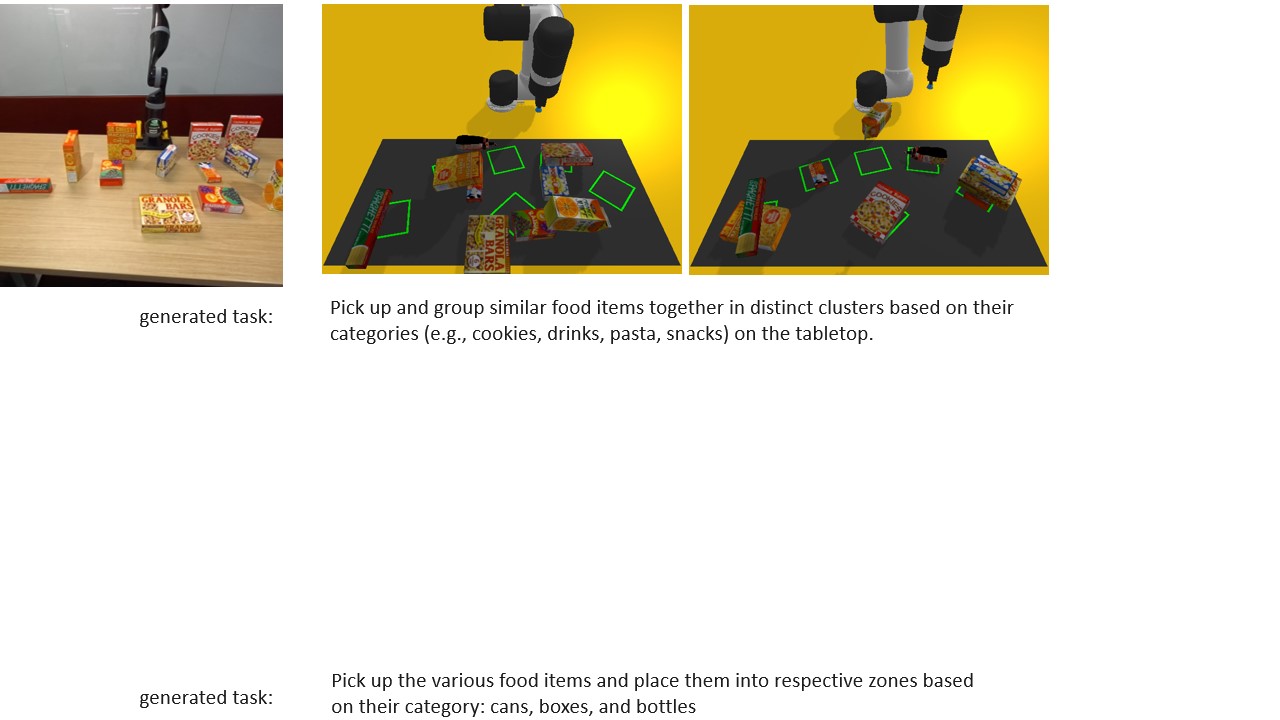}
    \caption{
    Qualitative results of five scenarios (rows) depicting task execution. Each row presents the input scene image (left), initial state (center), and final state after an oracle task execution (right). Task descriptions are provided beneath the central and rightmost columns. Note that task generation may not use all observed assets nor object orientations from the input image. Following GenSim we include the use of basic shapes like colored cubes and containers.
    }
    \label{fig:task_qual}
\end{figure}

We designed our simulation generation experiment to assess \sysname{}'s capacity to produce working simulators for robot policies.
Simulations use the CLIPort~\cite{shridhar2021cliport} task framework and we base our generation pipeline on the prompts used in GenSim~\cite{wang2023gensim} with minor modifications to indicate VLMs should use an input image.
GenSim prompts provide few-shot examples for task definitions and simulation code, along with API definitions that include a template base class for simulation code.
We evaluated task generation systems by assessing how well an oracle policy can complete the generated simulation tasks.
For evaluation scoring we used three runs of executing GenSim's oracle policy and averaged the results over these runs.

We compared our method ({\tt ours}) to three ablations:
1)~removing the \textit{router} and only fixing the simulation ({\tt no router}),
2)~removing the tests and only generating the simulation once ({\tt no fix}) and,
3)~removing the image input and using an LLM only for generating the simulation once ({\tt LLM})
The last ablation is the closest to the original GenSim~\cite{wang2023gensim}.
For simulation generation we only evaluated GPT4o given its superior performance in object correspondence.
For each of our 10 scenes we ran 30 generations and allowed up to 10 attempts to fix the code (either simulation or tests) in each run.
The same process was followed for all ablations, where {\tt no router} was limited to 10 test fixes and {\tt ours} was limited to a combined total of 10 fixes of either the simulation or tests.

\begin{table}[tb]
    \caption{Rewards obtained on generated tasks.}
    \centering
    {\small
\begin{tabular}{llllll}
\hline

ablation & reward \\
\hline
{\tt LLM}  & 0.47 \\
{\tt no fix} & 0.49 \\
{\tt no router} & 0.65 \\
{\tt ours} & \textbf{0.71} \\
\hline
\end{tabular}
}
    \label{tab:ablation_results}
\end{table}

Simulations generated by our method enabled the oracle policy to obtain higher rewards than alternatives, see Table~\ref{tab:ablation_results}.
We report average rewards over the oracle policy executions for each generated simulation over all 10 scenes.
As our interest is using these environments for robotic training we exclude cases with runtime errors.
All results are normalized where a reward of 1 indicates completion of all goals in the task.
We found our method yielded simulations that were effective for policy execution, and that removing test feedback ({\tt no fix} and {\tt LLM}) substantially lowered policy execution success.
Figure~\ref{fig:task_qual} presents qualitative results of simulations generated by our method.

We also examined the simulation and test fixing behavior to understand the router's behavior.
The router performed 0.52 fixes to tests on average across tasks, indicating there was a frequent need to repair the initially generated tests.
In addition, the router performed 1.08 fewer simulation fixes compared to {\tt no router} (5.81 vs 6.89), indicating the router was more efficient when making changes, making net 0.56 fewer total changes per generation.
This supports the claim that the router enabled more efficient automation of simulation generation.

\subsection{Qualitative Analysis of Code Generation}

To better understand \sysname{}'s behavior, we qualitatively examined the behavior of the router and the changes made when fixing the simulations and tests.

\noindent\textbf{Router. }
The router demonstrated a useful behavior of choosing to change the tests when the test feedback is too sparse to diagnose why the oracle fails.
When errors occurred, the router appropriately parsed error feedback to recognize the need to fix tests when there were problems with missing imports or cases where the test misused simulation objects, \textit{e.g.,} assuming an object was a list when it was not.

\noindent\textbf{Simulation Fixing. }
Often the simulation task was too complex to be successfully executed by the oracle.
We observed different behaviors to address this problem:
1)~simplifying the simulation by reducing the number of objects used and thus simplifying the task goals, 
2)~increasing the maximum number of steps the oracle could take before terminating an attempt, or 
3)~increasing the size of target areas for placing objects.
These fixes were typically in response to running tests where the oracle achieved only partial success on a task. 
This demonstrates that our proposed system was able to correctly parse nuanced results where no explicit error is raised beyond reporting that the oracle obtained a low reward.
However, on occasion this diverge from our intended outcome, where the fix removed the provided object spatial locations and only retained the assets.

\noindent\textbf{Test Fixing. } During test execution the oracle could fail the task but not produce meaningful errors due to the absence of simulator runtime errors.
In response, the test fixes added diagnostics about the oracle execution.
These monitored key performance indicators, including step count, intermediate goal achievement, and reward accumulation, thereby providing feedback signals throughout the testing process.
Moreover the test fixes extended beyond mere performance tracking, addressing fundamental issues such as erroneous interaction with simulation environment components and goal misinterpretation. 
The test fixes also focused on validating the correct initialization and reset functionality of the simulation environment. 
These measures ensured proper goal setup and the generation of valid observational data.
Note that these improvements were achieved while adhering to our principle of using generic prompts, which were designed solely to guide the system in testing oracle success.
This underscors the effectiveness of our refined test protocols in enhancing system evaluation without altering the core testing paradigm.
Further, this approach has promise for improving with better baseline language models, requiring no modifications to our framework.

\noindent\textbf{Common Failure Cases. }
During the experiments we discovered simulation code had rarely (roughly 30 times in 1200 generations) included while loops that could fail to terminate, \textit{e.g.,} looping indefinitely when trying to find a valid position to place assets in a constrained area.
This behavior did not always cause to failure or appear during testing, as most of the time the behavior would be correct.
The non-terminating loop condition meant the simulation execution would never exit, hanging the generation process.
This problem could be addressed by explicitly prompting to avoid using these kinds of loops and non-terminating behavior, or implementing timeout behavior in the baseline generation and testing framework.

During the fixing process the LLM sometimes misdiagnosed the cause of test failures and implemented new code as workarounds.
In simple cases the LLM would write geometric manipulation and other primitive functions instead of using the provided API, \textit{e.g.,} converting 3D rotations between quaternions and Euler angle representations, sampling from a probability distribution, inverting poses, \textit{etc.}
In rare cases this led to re-implementing the underlying reward functions or oracle agent in their entirety.
Sometimes the LLM would mock out the simulation environment creation or key aspects of the simulation behavior, bypassing the desired behavior altogether.
We anticipate improvements in LLM capabilities will yield better error diagnosis and code fixing over time, potentially addressing these limitations without requiring changes to our framework.

\subsection{Scene-level Extension}
\label{sec:scene-extension}

While our primary experiments focused on table-top scenes with a constrained asset set, we also tested our approach's scalability to more complex environments. 
We extended our pipeline to process scenes using approximately 150,000 assets from Objaverse~\cite{deitke2023objaverse}, significantly increasing the asset database size.

\begin{figure}[tb]
    \centering
    \includegraphics[width=0.45\linewidth]{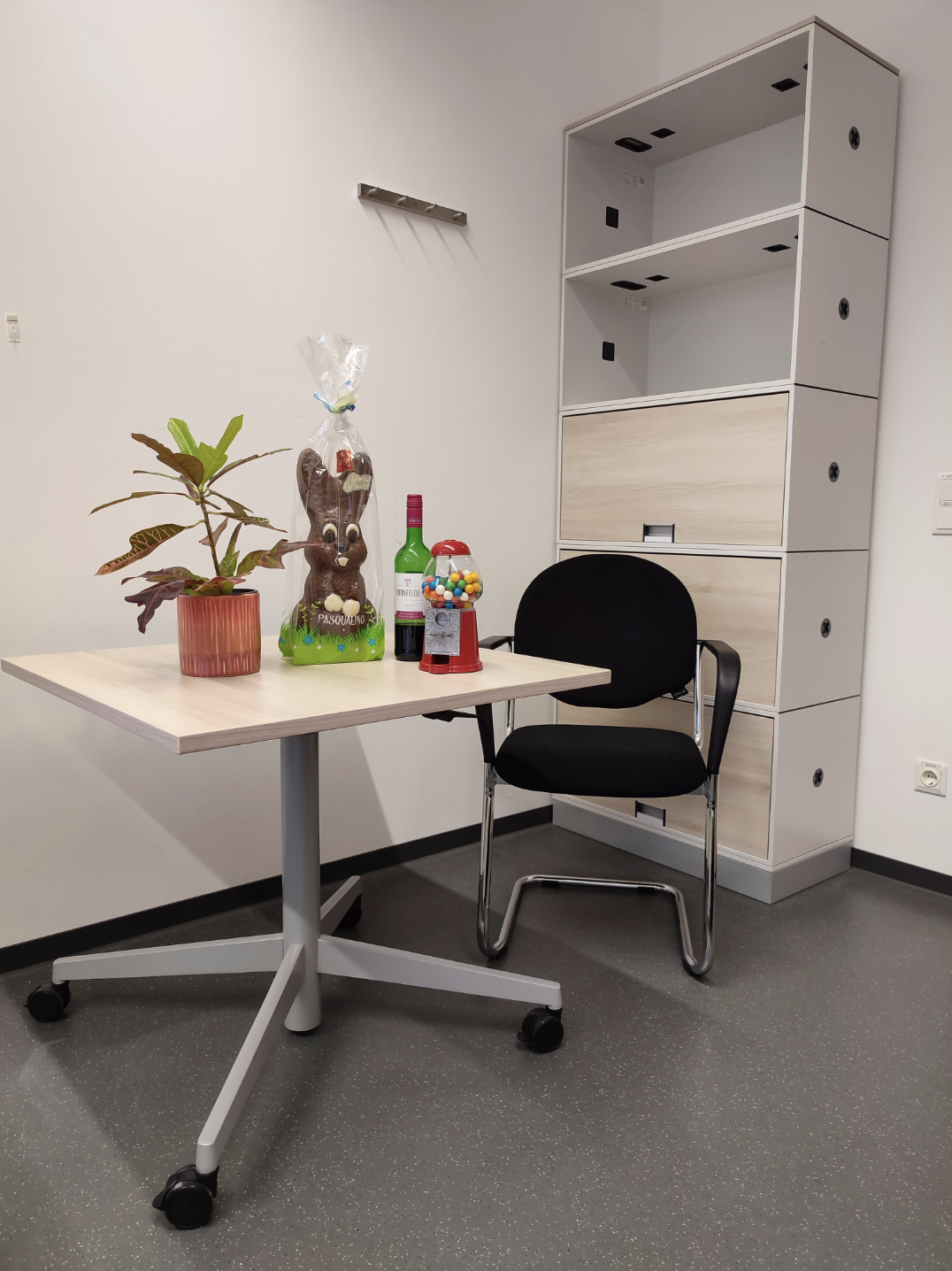}
    \includegraphics[width=0.45\linewidth]{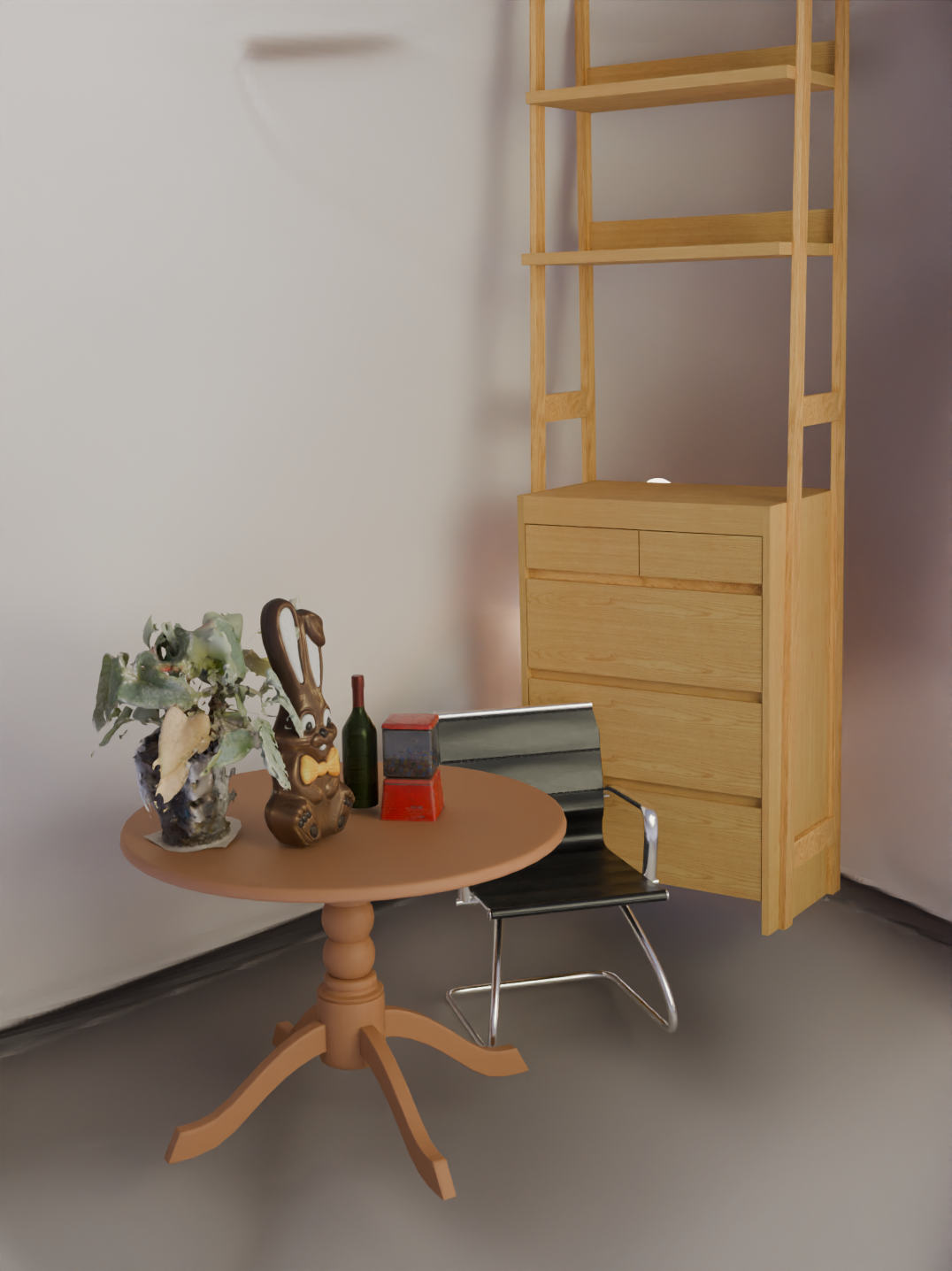}
    \caption{Real-world image (left) and a simulation environment in 3D (right) obtained with background reconstruction plus \sysname{}'s scene comprehension pipeline, using Objaverse~\cite{deitke2023objaverse} as the asset repository.}
    \label{fig:demo}
\end{figure}

Starting from a single RGB observation, we reconstructed the background of the scene by employing background estimation, fitting an MLP to estimate the SDF of the background surface, and then applying the marching cubes algorithm to generate the background mesh, following Dogaru \textit{et al.}~\cite{dogaru2024generalizable}. 
Once the background was reconstructed, we used \sysname{}'s real-to-sim workflow---combining object segmentation with VLM-based object matching---to construct the 3D task environment, see Figure~\ref{fig:demo}. 
The same object correspondence process described in Section~\ref{sec:scene-comprehension} was applied, but with the much larger Objaverse repository as the source of potential matching assets.

The results, as shown in Figure~\ref{fig:demo}, demonstrate that our approach can successfully scale to more complex scenes and significantly larger asset databases. 
This extension represents a preliminary step towards generating more complex scene-level tasks, where we leave further development of this direction to future work, as discussed in the following section.

\subsection{Limitations and Future Work}

Building on our experimental results and the scene-level extension, we identify several limitations and opportunities for future work.
Our experiments followed GenSim in assuming basic physics properties for scene assets. 
Future asset databases could provide richer metadata including mass, inertia, friction properties, and stiffness localized to asset parts.
Our method readily extends to retrieving such assets for appropriately equipped simulators. More ambitiously, foundation models could be used to reason about physics to design tasks targeted to specific objects' physical properties.
For the scene-level extension, further work is needed to generate semantically meaningful tasks for complex environments by extending our router to handle increased complexity and spatial relationships in full scenes beyond tabletop arrangements.

Future work could also explore sim-to-real training and transfer learning techniques. Integrating \sysname{} with sim-to-real processes could leverage the router to refine automatically revise the simulation and tests to close the sim-to-real gap.
Using differentiable physics engines like Newton~\cite{nvidia2025newton} could accelerate agent training through the simulator for reinforcement learning.
Combined with our router, this could streamline the sim-to-real pipeline by automating both simulation generation and policy training.

\section{Conclusions}

In this work, we presented \sysname{}, a system for generating robotic simulation tasks from real-world observations.
Our approach integrates scene understanding with VLMs, asset matching, and task generation through an innovative dual-generation process and router system.
Experiments demonstrate effective real-to-sim translation from single RGB-D observations and scalability to more complex scenes.

The implications of this work extend beyond robotics to game development, where \sysname{} can automate realistic environment creation, generate solvable game tasks, and reduce development cycles for applications requiring realistic physics interactions.
This approach is especially valuable for serious games and educational applications where real-world fidelity affects learning outcomes.
Future work could address more complex scenes, improve asset matching, integrate physics-based reasoning, and explore transfer learning techniques as we continue developing digital twins across application domains.





{
    \small
    \bibliographystyle{ieeenat_fullname}
    \bibliography{bib}
}

\section{Appendix}

\subsection{Example Task Definition}

Below we provide full code listings for a generated task definition, simulation, and test that result from running \sysname{} on a real-world image.

\begin{lstlisting}[caption=Example task definition.]
    "task-name": "organize-food-items",
    "task-description": "Pick up all the food items on the table and place them into the open box in the center.",
    "assets-used": [
        "HOPE/GranolaBars.urdf",
        "HOPE/MacaroniAndCheese.urdf",
        "HOPE/OrangeJuice.urdf",
        "HOPE/Raisins.urdf",
        "HOPE/Mustard.urdf",
        "HOPE/Cookies.urdf"
    ]
\end{lstlisting}

\begin{lstlisting}[language=Python, label=lst:task-code, caption=Example task code.]
class PackFoodItemsInBox(Task):
    """Pick up all the food items and place them inside the open box."""

    def __init__(self):
        super().__init__()
        self.max_steps = 20  # Increased max steps to give the oracle more time
        self.lang_template = "pack the {} in the brown box"
        self.task_completed_desc = "task completed."
        self.sixdof = True  # Allowing six degrees of freedom for more accurate placement
        self.additional_reset()

    def reset(self, env):
        super().reset(env)

        # Object bounding boxes.
        object_bounding_boxes = [
            {'bbox_corners': [np.array([0.76183139, -0.0314516, -0.00410459]), np.array([0.62986134, 0.13964655, 0.0253295])], 'urdf': 'HOPE/GranolaBars.urdf'}, 
            {'bbox_corners': [np.array([0.51239947, 0.20819839, 0.01941574]), np.array([0.2688748, 0.42548292, 0.21593083])], 'urdf': 'HOPE/MacaroniAndCheese.urdf'}, 
            {'bbox_corners': [np.array([0.53965911, -0.01996892, 0.05628724]), np.array([0.45665461, 0.05610127, 0.24530269])], 'urdf': 'HOPE/OrangeJuice.urdf'}, 
            {'bbox_corners': [np.array([0.43465213, -0.16156002, 0.02437929]), np.array([0.378738, -0.08377419, 0.14193851])], 'urdf': 'HOPE/Raisins.urdf'}, 
            {'bbox_corners': [np.array([0.53083534, 0.16836134, -0.02701451]), np.array([0.4384108, 0.31221087, 0.0206661])], 'urdf': 'HOPE/Mayo.urdf'}, 
            {'bbox_corners': [np.array([0.6035398, 0.10533048, -0.01784448]), np.array([0.52428974, 0.26218855, 0.03146984])], 'urdf': 'HOPE/Mustard.urdf'}, 
            {'bbox_corners': [np.array([0.64027918, -0.45123034, 0.05872372]), np.array([0.48848391, -0.34876966, 0.23379576])], 'urdf': 'HOPE/Cookies.urdf'}
        ]

        # Add the box.
        box_size = (0.3, 0.2, 0.1)
        box_pose = ((0.5, 0, box_size[2] / 2), (0, 0, 0, 1))  # Ensure the box is centered and above the table
        box_template = 'container/container-template.urdf'
        replace = {'DIM': box_size, 'HALF': (box_size[0] / 2, box_size[1] / 2, box_size[2] / 2)}
        box_urdf = self.fill_template(box_template, replace)
        box_id = env.add_object(box_urdf, box_pose, 'fixed')

        # Add food items.
        objects = []
        for bbox in object_bounding_boxes:
            urdf = bbox['urdf']
            corners = bbox['bbox_corners']
            obj_size = corners[1] - corners[0]
            
            # Ensure the object is placed randomly within the bounds and above the table
            obj_pose, obj_rot = self.get_random_pose(env, obj_size)
            while obj_pose is None:
                obj_pose, obj_rot = self.get_random_pose(env, obj_size)
            
            obj_id = env.add_object(urdf, (obj_pose, obj_rot))
            objects.append(obj_id)

        # Ensure the object poses are above the table and within bounds
        valid_targ_poses = []
        for obj_id in objects:
            pos, _ = p.getBasePositionAndOrientation(obj_id)
            if (self.bounds[0, 0] <= pos[0] <= self.bounds[0, 1] and
                self.bounds[1, 0] <= pos[1] <= self.bounds[1, 1] and
                self.bounds[2, 0] <= pos[2] <= self.bounds[2, 1]):
                valid_targ_poses.append((pos, (0, 0, 0, 1)))

        # Define the goals.
        for i, obj_id in enumerate(objects):
            self.add_goal(
                objs=[obj_id], 
                matches=np.int32([[1]]), 
                targ_poses=[box_pose], 
                replace=False,
                rotations=True, 
                metric='pose',  # Changed to 'pose' for simpler task completion
                params=[], 
                step_max_reward=1 / len(objects),
                language_goal=self.lang_template.format(object_bounding_boxes[i]['urdf'].split('/')[1].split('.')[0])
            )
\end{lstlisting}

\begin{lstlisting}[language=Python, caption=Example test code.]
class TestPackFoodItemsInBox(unittest.TestCase):

    def setUp(self):
        assets_root = os.path.join(os.environ.get('CLIPORT_PATH', os.getcwd()), "cliport/environments/assets/")
        self.env = Environment(
            assets_root,
            disp=False,
            hz=480,
            record_cfg=[],
        )
        self.task = PackFoodItemsInBox()
        self.env.set_task(self.task)
        self.env.reset()

    def test_oracle_succeeds(self):
        oracle_agent = self.task.oracle(self.env)
        obs = self.env.reset()
        info = self.env.info

        total_reward = 0
        for _ in range(self.task.max_steps):
            act = oracle_agent.act(obs, info)
            if act is None:
                break  # Ensure we handle cases where the oracle cannot generate an action
            obs, reward, done, info = self.env.step(act)
            total_reward += reward
            
            if done:
                break
        
        # Ensure that the task is completed successfully.
        self.assertTrue(done, "Oracle agent did not complete the task within the allowed steps.")
        self.assertAlmostEqual(total_reward, 1.0, delta=0.01, msg="Oracle agent did not achieve maximum reward.")

    def test_oracle_picks_and_places_objects(self):
        oracle_agent = self.task.oracle(self.env)
        obs = self.env.reset()
        info = self.env.info

        pick_count = 0
        place_count = 0

        for _ in range(self.task.max_steps):
            act = oracle_agent.act(obs, info)
            if act is None:
                break  # Ensure we handle cases where the oracle cannot generate an action
            obs, reward, done, info = self.env.step(act)
            
            if act and 'pose0' in act and 'pose1' in act:
                pick_count += 1
                place_count += 1
            
            if done:
                break
        
        # Ensure that the oracle agent attempted to pick and place objects.
        self.assertGreater(pick_count, 0, "Oracle agent did not attempt to pick any objects.")
        self.assertGreater(place_count, 0, "Oracle agent did not attempt to place any objects.")

    def test_task_completion_description(self):
        oracle_agent = self.task.oracle(self.env)
        obs = self.env.reset()
        info = self.env.info

        total_reward = 0
        for _ in range(self.task.max_steps):
            act = oracle_agent.act(obs, info)
            if act is None:
                break  # Ensure we handle cases where the oracle cannot generate an action
            obs, reward, done, info = self.env.step(act)
            total_reward += reward
            
            if done:
                break
        
        # Ensure that the task completion description is accurate.
        self.assertTrue(done, "Oracle agent did not complete the task within the allowed steps.")
        self.assertEqual(self.task.task_completed_desc, "task completed.", "Task completion description does not match.")

if __name__ == '__main__':
    unittest.main()
\end{lstlisting}

\subsection{Prompts}

Below we provide the prompts used for the oracle test generation, including the prompt and example of the oracle API.

\begin{lstlisting}[language=Python, caption=Prompt for behavior test.]
"""You are an AI in robot simulation code and task design.
My goal is to design diverse and feasible tasks for tabletop manipulation.

Write tests to ensure a robotic policy could solve this task.
Do not write tests for the environment set or goal structures, only the behavior of the robot policy itself.
The task is defined is in a file named "task_code.py". Only import the task from that file.

Provide your response as a fully runnable python unittest script.
Only provide the script content; do not provide instructions on how to run it.
Do NOT mock ANY inputs to the test.
Use the guidance below to set up and initialize the environment for tests.

Here is the task definition:
{task}

Here is task code:
```python
{code}
```"""
\end{lstlisting}

\begin{lstlisting}[language=Python, caption=Oracle prompt code.]
"""Here is how to run the oracle agent on a task (`task` subclassing `Task`) in an environment (`env` of type `Environment`):

```
oracle_agent = task.oracle(env)
obs = env.reset()
info = env.info

total_reward = 0
for _ in range(task.max_steps):
    act = oracle_agent.act(obs, info)
    obs, reward, done, info = env.step(act)
    total_reward += reward
    
    if done:
        break
```"""
\end{lstlisting}

\end{document}